%% file: main.tex
\documentclass[10pt,twocolumn,letterpaper]{article}

\usepackage[pagenumbers]{cvpr} 

\usepackage{booktabs}
\usepackage{graphicx}
\usepackage{nicefrac}
\usepackage{xspace}

\newcommand{\testpub}{\textit{TEST}$_{\text{pub}}$\xspace}
\newcommand{\testpriv}{\textit{TEST}$_{\text{priv}}$\xspace}
\newcommand{\testprivmix}{\textit{TEST}$_{\text{priv,mix}}$\xspace}
\newcommand{\fone}{$F_1$\xspace}
\newcommand{\percentile}{th percentile\xspace}

\definecolor{cvprblue}{rgb}{0.21,0.49,0.74}
\usepackage[pagebackref,breaklinks,colorlinks,allcolors=cvprblue]{hyperref}

\def\paperID{*****} 
\def\confName{CVPR}
\def\confYear{2026}

\title{SuperADD: Training-free Class-agnostic Anomaly Segmentation -- CVPR 2026 VAND 4.0 Workshop Challenge Industrial Track}

\author{
\begin{tabular}{@{}cccc@{}}
Lukas Roming$^{1}$ & Felix Lehnerer$^{1}$ & Jonas V.~Funk$^{2}$ & Andreas Michel$^{1}$ \\
\end{tabular} \\[6pt]
\begin{tabular}{@{}ccc@{}}
Georg Maier$^{1}$ & Thomas L\"angle$^{1}$ & J\"urgen Beyerer$^{1}$ \\
\end{tabular} \\[4pt]
\\
$^1$Fraunhofer IOSB, Fraunhoferstr.~1, 76131 Karlsruhe, Germany \\ \quad 
$^2$Independent Research\\[1pt]
{\tt\small lukas.roming@iosb.fraunhofer.de} \\
}

\begin{document}
\maketitle
\input{sec/0_abstract}
\input{sec/1_intro}

\input{sec/2_Methodology}
\input{sec/3_Results}
\input{sec/4_discussion_and_conclusion}
\newpage
{
    \small
    \bibliographystyle{unsrt} 
    \bibliography{main}
}


\end{document}

%% file: sec/0_abstract.tex
\begin{abstract}
Visual anomaly detection (AD) for industrial inspection is a highly relevant task in modern production environments. The problem becomes particularly challenging when training and deployment data differ due to changes in acquisition conditions during production. In the \emph{VAND 4.0} Industrial Track, models must remain robust under distribution shifts such as varying illumination and their performance is assessed on the MVTec AD 2 dataset.
To address this setting, we propose a training-free and class-agnostic anomaly detection pipeline based on the work of SuperAD. Our approach improves generalization through several modifications designed to enhance robustness under distribution shifts. These adaptations include using a DINOv3 backbone, overlapping patch-wise processing, intensity-based augmentations, improved memory-bank subsampling for better coverage of the data distribution, and iterative morphological closing for cleaner and more spatially consistent anomaly maps. Unlike methods that rely on class-specific architectures or per-class hyperparameter tuning, our method uses a single architecture and one shared hyperparameter configuration across all object classes. This makes the approach well suited for industrial deployment, where product variants and appearance changes must be handled with minimal adaptation effort. We achieve segmentation \fone scores of \ $62.61\%$, $57.42\%$, and $54.35\%$ on \testpub, \testpriv, and \testprivmix of MVTec AD 2, respectively, thereby surpassing SuperAD and other state-of-the-art methods. Code is available at \url{https://github.com/LukasRoom/SuperADD}.
\end{abstract}

%% file: sec/1_intro.tex
\section{Introduction}
\label{sec:intro}

\subsection{Background}
\label{subsec:background}

Automated visual anomaly detection (AD) is increasingly used in industrial inspection, where high-throughput production and subtle defects create demanding operating conditions. Despite this growing adoption, performance often degrades under varying acquisition conditions, particularly illumination shifts that alter object appearance without affecting semantic structure~\cite{li2025survey}. This sensitivity motivates the need for the \emph{VAND 4.0} Industrial Track, which focuses on robustness to such variations while maintaining real-time efficiency.

\subsection{Challenge Description}
\label{subsec:challenge-description}

The \emph{VAND 4.0} Industrial Track evaluates AD models under acquisition shifts. Models are trained exclusively on normal images and must detect anomalies at test time. The task emphasizes balancing detection performance with computational efficiency.

\textbf{Dataset:} The challenge uses the MVTec Anomaly Detection 2 (MVTec AD 2) dataset~\cite{heckler2026}, a benchmark for unsupervised AD~\cite{heckler2025}. The dataset comprises eight real-world classes captured under varying lighting conditions, leading to appearance shifts between training and test data. The absence of openly available ground truth for the private test split further increases its resemblance to real-world conditions by enforcing blind evaluation and reducing the risk of overfitting. The eight object classes can be grouped according to visual characteristics (non-exclusive):
\begin{itemize}
    \item Bulk goods objects: \textit{wall plugs}, \textit{walnuts}, and \textit{rice}, featuring overlapping and occluded instances arranged in uncontrolled spatial patterns.
    \item Textured objects: \textit{fabric} and \textit{sheet metal}, exhibiting high variability in normal appearance.
    \item Reflective metal objects: \textit{sheet metal} and \textit{can}.
    \item Transparent objects: \textit{vial} and \textit{fruit jelly}, which are particularly sensitive to illumination changes.
\end{itemize}

In addition, a range of lighting conditions is applied to the private split (\testpriv), resulting in an additional split further amplifying distribution shifts between training and test data (\testprivmix). 

\textbf{Challenge Goal:} Methods are evaluated by their pixel-level \fone score and ranked separately on \testpriv and \testprivmix of MVTec AD~2. The final score is the mean rank across these two test sets.
In contrast to \emph{VAND 3.0}, where successful methods often relied on class-specific architectures or class-specific hyperparameter tuning~\cite{heckler2025}, this year's focus is on a single class-agnostic solution that generalizes across diverse objects. To enforce this, proposed solutions must use the same architecture and hyperparameters across classes.

\subsection{Related Work}
\label{subsec:related-work}
The original MVTec AD benchmark~\cite{bergmann2019} has become largely saturated, with state-of-the-art methods achieving very high segmentation performance and leaving little room for meaningful comparison. This limits its usefulness for distinguishing between recent AD approaches and motivates the development of more challenging benchmarks such as MVTec AD 2.

Among the wide range of unsupervised AD methods, the strongest solutions in last year’s \emph{VAND 3.0} challenge (ISVL~\cite{wang2025}, RoBiS~\cite{li2025}) were largely based on INP-Former~\cite{luo2025}. INP-Former is a reconstruction-based AD method that requires training to extract intrinsic normal prototypes from the input image via a Vision Transformer and reconstructs only the normal content. Regions that cannot be reconstructed well are treated as anomalies, and the reconstruction residual provides the anomaly score. 

In contrast, we follow the memory bank approach of SuperAD~\cite{zhang2025superad}, inspired by PatchCore~\cite{roth2022}, adopting a training-free strategy that avoids model retraining. In SuperAD, normal images are used to build a memory bank using the feature extraction capabilities of DINOv2~\cite{oquab2023dinov2}.

We build on this architecture because its training-free design is easily adaptable and provides an attractive alternative to methods that require dedicated retraining. This makes the method particularly appealing for industrial deployment, where new product classes or design changes often need to be integrated quickly.

\subsection{Our Contributions}
\label{subsec:our-contributions}
While SuperAD already performed strongly in last year's challenge, we focus on improving the underlying architecture through generalizable modifications. Our goal is a robust and practically deployable AD pipeline that remains effective across diverse object classes. Our main contributions are as follows:
\begin{itemize}
    \item We maintain a class-agnostic setup, using the same architecture and hyperparameters across all object classes.
    \item We improve the subsampling stage in memory bank construction for computational efficiency and better data distribution coverage.
    \item We introduce overlapping patch-wise preprocessing to reduce grid-position sensitivity and improve generalization.
    \item We apply a simple iterative morphological closing step in postprocessing to obtain more spatially consistent anomaly maps.
    \item We employ intensity-based augmentations to simulate varying illumination conditions, improving robustness to lighting changes between training and test data.
    \item We replace the original DINOv2 backbone with DINOv3, leveraging improved pre-trained visual representations.
\end{itemize}

%% file: sec/2_Methodology.tex
\section{Methodology}
\label{sec:methodology}

\subsection{Model Design}
\label{subsec:model-design}

\subsubsection{Approach}
\label{subsubsec:approach}


We build upon SuperAD from Zhang et al.~\cite{zhang2025superad}. We chose this architecture for its training-free and flexible design enabling faster development, hyperparameter tuning, and deployment.
Just like SuperAD~\cite{zhang2025superad} and PatchCore~\cite{roth2022}, we create a memory bank at multiple intermediate layer outputs of a powerful visual feature extractor and perform nearest-neighbor-based outlier-detection at inference. As feature extractor, we employ DINOv3~\cite{simeoni2025dinov3} due to its state-of-the-art performance on a range of downstream classification and segmentation benchmarks.

To improve computational efficiency, we adopt a refined subsampling strategy. SuperAD applies PatchCore’s coreset-reduction method to class tokens to select 16 images from which to extract memory-bank feature vectors. While this promotes diversity at the image level, it does not effectively eliminate near-duplicate feature vectors within the same image or similar regions appearing in different images, such as those arising from homogeneous background regions or from multiple, low-variation instances of the same object.

To address this limitation, we perform subsampling directly on the feature vectors rather than on the images. Specifically, we use a nearest-neighbor-based subsampling procedure with the same objective as the coreset-reduction approach of PatchCore, to remove near-duplicate embeddings from the memory bank. Details on the subsampling are given in section \ref{subsubsec:training}.

Due to the position invariance of several classes (\textit{rice}, \textit{walnuts}, and \textit{wall plugs}), we adopted patch-wise image processing to reduce position-dependent artifacts. These artifacts include false AD in empty regions that the model was not trained on. Patch-wise inference limits the model’s global view and thus its sensitivity to such cases, as seen for wall plugs. Patch-wise processing improved performance for most classes (except fabric and vial).

To enhance the robustness of AD, we further apply intensity scaling to the training images to simulate variations in integration time.

In the original SuperAD implementation, class-specific thresholds are derived from test data with its ground truth by selecting the threshold that maximizes \fone score. Instead, we define the threshold as a scaled version of the $95$\percentile of the anomaly map values observed in the training data. This improves practical usability and complies with the \emph{VAND 4.0} guidelines.

\subsubsection{Architecture}
\label{subsubsec:architecture}

An overview of the proposed method is illustrated in Figure \ref{fig:architecture}. The main components are patching, a vision transformer backbone, and k-nearest neighbors comparison.

\textbf{Overlapping patches:}
The input image of shape $H \times W$ is divided into equally sized patches of size $P \times P$. The number of patches $n_d$ along a dimension $d$ of size $L_d$ with a minimum overlap of size $O$ is then determined by \eqref{eq:n_patches_dim}. The start positions of patches are then spread linearly between $0$ and $L_d - P$ such that the first patch starts at $0$ and the last patch ends at $L_d - 1$. This guarantees a minimum overlap between patch borders inside the image. The overlap between patches reduces the vulnerability to artifacts occurring at the edges of the inference window. The alignment with image borders eliminates the need for padding which would lead to unrealistic reference embeddings. For example, zero padding could result in incomplete objects being labeled as non-anomalous. We chose a patch size of $P=640$ with a minimum overlap of $O=128$.

\begin{equation}
\label{eq:n_patches_dim}
    n_{d}
    = \left\lceil \frac{L_d - P}{P - 2O} \right\rceil + 1,
\end{equation}

\textbf{Backbone:}
We adopt the pretrained model DINOv3-ViT-H+/16 consisting of 32 transformer layers (or blocks) with approx. 840 million parameters. We extract embeddings from the layers 7, 15, 23, and 31. After inference, redundant predictions in the overlapping regions at patch borders are discarded. The remaining embedding vectors are then mapped back to their original spatial positions and concatenated, producing a single coherent inference result for the whole image.

\textbf{Deriving anomaly maps:}
Following SuperAD~\cite{zhang2025superad}, we perform nearest-neighbor search for the selected layer outputs and average across the nearest-neighbor distance maps. Finally, we post-process the floating point anomaly maps using thresholding, closing, and a fill region method as described in section \ref{subsec:postprocessing}.

\definecolor{train_green}{RGB}{0,150,0}
\definecolor{test_blue}{RGB}{0,0,150}

\begin{figure*}
    \centering
    \includegraphics[width=1\linewidth]{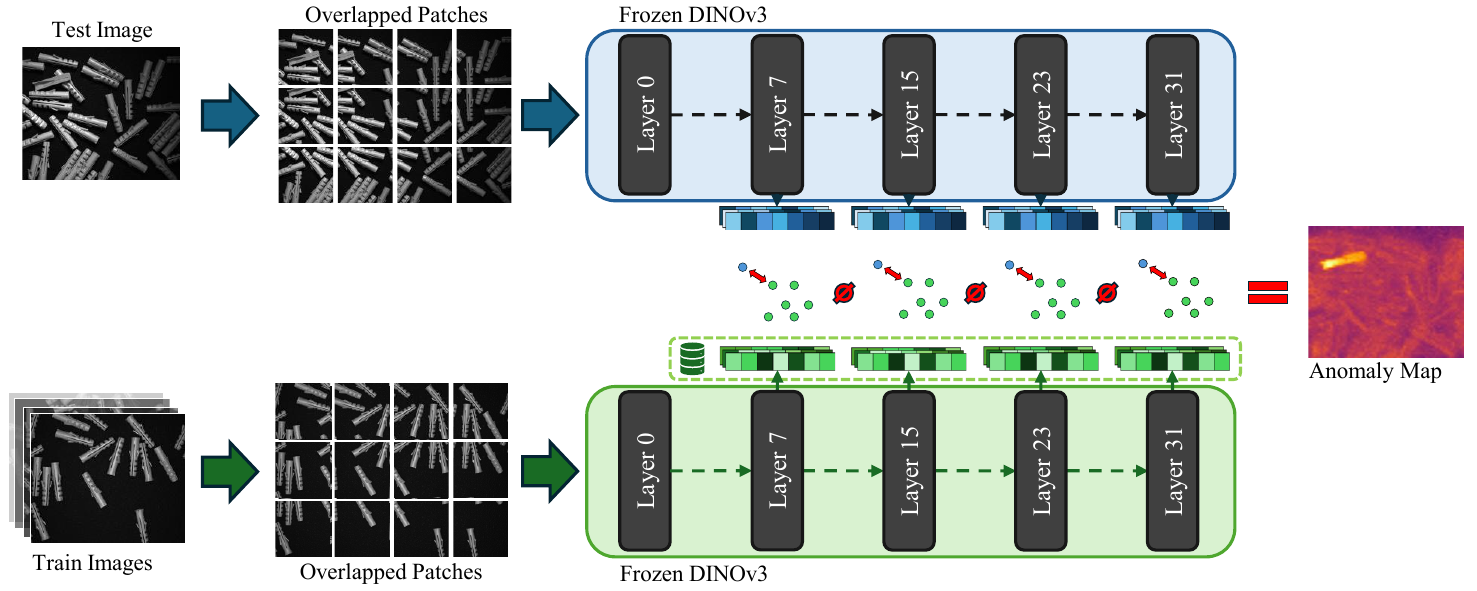}
    \caption{Overview of the proposed method to calculate anomaly maps based on SuperAD~\cite{zhang2025superad}. Inference is performed on patches for \textcolor{test_blue}{test} and \textcolor{train_green}{train} images. Embeddings are extracted from multiple layers to create prototypes representing anomaly-free data. The anomaly score is determined by the average distance between a test embedding vector to its nearest neighbor prototype.}
    \label{fig:architecture}
\end{figure*}


\subsubsection{Training}
\label{subsubsec:training}

Just like SuperAD, our method does not have any parameters to be trained and all model weights stay fixed. We use the training data to extract prototype embeddings from four layers (as described in section \ref{subsubsec:architecture}) and derive an anomaly threshold.




\textbf{Pre-processing:}
First, images are downscaled by a factor of $0.625$ and normalized using ImageNet normalization (this is also used for inference). Normalized pixel values are then scaled by a random factor uniformly sampled from the interval $[0.8, 1.2]$. This is done to improve robustness to the distribution shift caused by varying lighting conditions.

\textbf{Prototype embedding subsampling:}
Instead of selecting 16 images of the training samples for prototype generation as suggested by~\cite{zhang2025superad}, we use all available training images (reserving $\nicefrac{1}{8}$ of them for threshold estimation) and apply a fast subsampling scheme to construct a compact and diverse memory bank. This reduces memory usage and inference time while removing redundant prototypes.

Specifically, we perform a $k$-nearest-neighbor-based selection in feature space:

\begin{enumerate}
    \item For each candidate feature vector, compute the distances to its $k=100$ nearest neighbors.
    \item Compute a global distance threshold $\tau$ as the mean of all $k$-NN distances.
    \item For each feature vector, define a subsampling score as the number of its $k$ neighbors whose distance is smaller than $\tau$. Vectors with low scores lie in sparsely populated regions of the feature space and are therefore more informative.
    \item Sort all feature vectors by this subsampling score in ascending order and retain the first $n$ vectors, where $n$ is the target memory-bank size.
\end{enumerate}

To further accelerate the procedure, it can be applied independently to several (random) subsets of the original feature set. The selected prototypes from all subsets are then merged. Empirically, this subset-wise subsampling not only reduces computation time but can also yield equal or slightly improved mean \fone scores compared to running the algorithm once on the full feature set.

\textbf{Threshold-estimation:} We process $\nicefrac{1}{8}$ of the training samples (that were not used for prototype generation) the same way as we do the inference of test samples, except that we stop at the floating point valued distance map. We therefore, compare those separated training samples to the prototype embeddings in the memory bank and calculate the nearest-neighbor distance map. Using the values of the map, we define the threshold such that it splits the $95$\percentile. We further scale this threshold by a constant gain factor, whereby a gain in the range of $1.3$ to $1.5$ worked well in our experiments. We found that his strategy is more robust and yields better results than using a higher percentile (e.g. $99$\percentile) without additional scaling.
\subsubsection{Inference}
\label{subsubsec:inference}

For inference of test images, we use the same model to compute embeddings and measure the distance to the nearest neighbor in the prototype memory bank. Distances are computed per 16×16 patch, corresponding to the model’s internal token size (not to be confused with the 640×640 patches used at a higher level of the architecture). Following PatchCore, this procedure is applied to multiple intermediate DINOv3 layers: for each of the four selected layers, we construct a memory bank, compute a distance map for the test image, and then average the resulting maps.

The final anomaly map has a spatial resolution of $\nicefrac{10}{256}$ of the original input image. This map is then upscaled to $\nicefrac{1}{4}$ of the original resolution and subsequently post-processed.


\subsubsection{Post-processing}
\label{subsec:postprocessing}
\begin{figure*}
    \centering
    \includegraphics[width=0.9\linewidth]{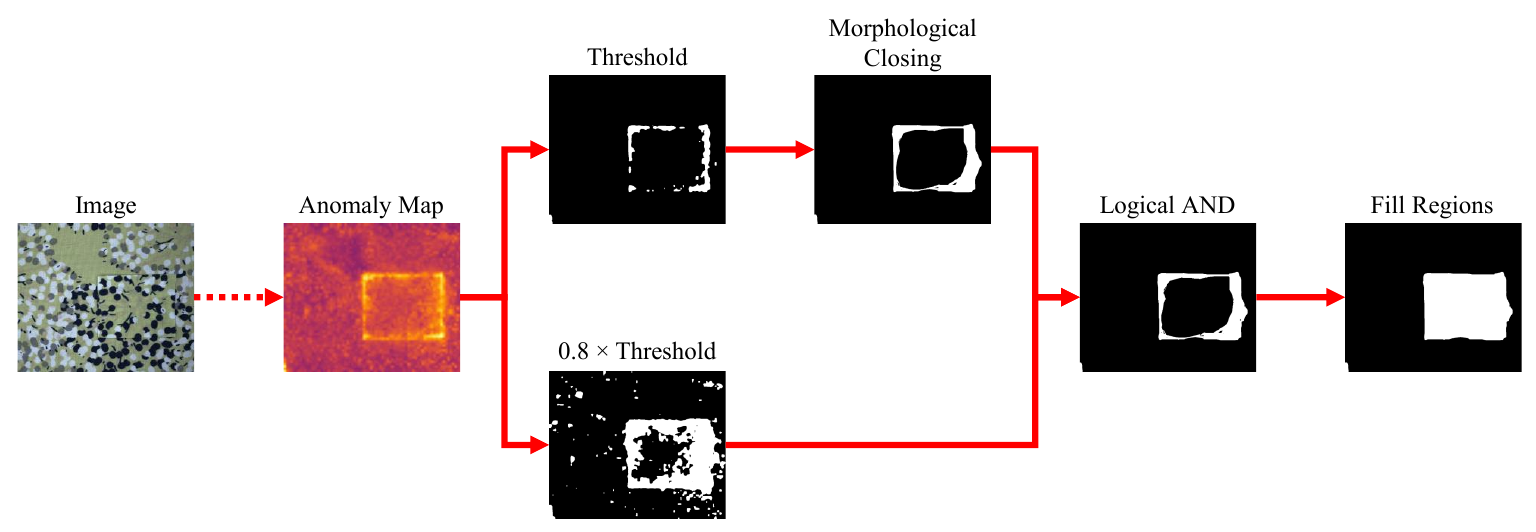}
    \caption{Overview of the postprocessing procedure. After initial thresholding, morphological closing is applied. The output is masked using a logical AND operation. Finally, closed regions are filled.}
    \label{fig:postprocessing}
\end{figure*}

Figure \ref{fig:postprocessing} shows an overview of the applied post-processing procedure. After the distance based anomaly scores are obtained a thresholding operation is applied with an automatically determined threshold value.

Morphological closing is applied to close small gaps in the detected binary masks to compensate for small false-negative detections at object or anomaly borders. A multi-oriented linear closing approach is applied to connect any broken linear features such as edges of anomalies. Morphological closing is performed 16 times using line structuring elements with a radius of 26 pixels and evenly spaced orientations between 0° and 180°. The results of the individual closing operations are combined with a pixelwise logical OR operation. To suppress unwanted artifacts the input image is zero-padded with $26+1$ pixels, and the padding is removed afterwards.

The result of morphological closing is masked using a binary mask obtained by thresholding the anomaly map at $0.8 \times$ the determined anomaly threshold, to prevent oversegmentation.

Due to the preceding patching step, the resulting binary mask may contain holes where anomalies extend across patch boundaries. In a final postprocessing step closed regions are filled to eliminate any holes. 

\subsection{Dataset \& Evaluation}
\label{subsec:dataset-eval}

\subsubsection{Dataset Utilization}
\label{subsubsec:dataset-utilization}

The MVTec AD 2 test data is divided into three subsets: \testpub, \testpriv, and \testprivmix . Ground-truth annotations are only available for \testpub, on which we perform local benchmarking. For \testpriv and \testprivmix, only the images are publicly provided and final scores were obtained via the official evaluation server. \testpriv contains images captured under the same lighting conditions as the training set, whereas \testprivmix  shows the same scenes but with a mixture of seen and unseen lighting conditions. For details about preprocessing and augmentations refer to section \ref{subsubsec:architecture}. The usage of training data is explained in section \ref{subsubsec:training}.


\subsubsection{Evaluation Criteria}
\label{subsubsec:evaluation-criteria}
Results of the described method are evaluated following the challenge guidelines. We therefore report the pixelwise \fone score \eqref{eq:f1_score} on the public test dataset as it best aligns with the rating criteria of the challenge. Additionally the $\text{AU-ROC}_{0.05}$ scores are provided. Obtained scores are solely used to assess the quality of detection but not used to automatically determine optimal class-specific thresholds.

\begin{equation}
\label{eq:f1_score}
F_1 = 2 \cdot \frac{\text{precision} \cdot \text{recall}}{\text{precision} + \text{recall}}
\end{equation}

\subsubsection{Evaluation Setting}
\label{subsubsec:evaluation-setting}
Our work focuses on the regular setting of the industrial track. For each category, we adapt a separate model instance using only the official anomaly-free training split. All model hyperparameters and thresholds are either static or selected automatically from training data in accordance with the challenge rules. The complete implementation, including all dependencies and a step-by-step guide to reproduce our results, is available at \url{https://github.com/LukasRoom/SuperADD}.


%% file: sec/3_Results.tex
\section{Results}
\label{sec:results}

\subsection{Quantitative Results}
\label{subsec:quantitative-results}

We report the \fone score and AU-ROC$_{0.05}$ for \testpub in Table \ref{tab:scores_test_private}.
We achieved high AU-ROC and \fone scores for all classes except \textit{can}. Samples of \textit{can} in \testpub contained fine defects that are barely visible to the human eye.
The method achieved the highest \fone score for \textit{fabric} ($93.74$\%), followed by \textit{wallplugs} ($79.16$\%). 

Further, we report the \fone scores for \testpriv and \testprivmix in Table \ref{tab:scores_test_private}. The highest F1-Scores of our approach for \testpriv were achieved for \textit{fabric} ($88.47$\%), matching the observation in \testpub, followed by \textit{rice} ($73.83$\%).
Interestingly, performance on \textit{wallplugs} is here significantly lower compared to \testpub, which can be attributed to the presence of substantially more difficult wallplug samples in \testpriv, characterized by more subtle defects and likely a reduced tolerance for false positives due to the low number of positive instances in the ground truth.

We compare our results in Table \ref{tab:scores_test_private} to the results of PatchCore, EfficientAD, and the top four approaches of last years challenge for \testpriv and \testprivmix. We achieved top \fone scores for several categories, including \textit{rice} and \textit{vial}. Most importantly, we outperformed all other methods in terms of mean \fone score. We achieved $57.42$\% and $54.35$\% for \testpriv and \testprivmix respectively, compared to last years best (ISVL~\cite{wang2025}) with $53.81$\% and $51.43$\% .


\begin{table}[h]
\centering
\begin{tabular}{lcc}
\toprule
\textbf{Object} & $\mathbf{AU\text{-}ROC_{0.05}}$ & \textbf{\fone score} \\
\midrule
Can          & 51.61 & 0.00 \\
Fabric       & 84.20 & 93.74 \\
Fruit Jelly  & 81.19 & 54.68 \\
Rice         & 96.11 & 73.31 \\
Sheet Metal  & 93.18 & 59.54 \\
Vial         & 82.59 & 64.77 \\
Wallplugs    & 92.53 & 79.16 \\
Walnuts      & 90.03 & 75.69 \\
\midrule
Mean         & 83.93 & 62.61 \\
\bottomrule
\end{tabular}
\caption{Segmentation AU-ROC$_{0.05}$ and \fone score (in \%) for the proposed method on binarized images for \testpub.}
\label{tab:scores_test_public}
\end{table}


\begin{table*}[t]
\centering
\resizebox{\linewidth}{!}{%
\begin{tabular}{lccccccc}
\toprule
\textbf{Category} & \textbf{PatchCore~\cite{roth2022}} & \textbf{EfficientAD~\cite{batzner2024}} & \textbf{ISVL~\cite{wang2025}} & \textbf{RoBiS~\cite{li2025}} & \textbf{ASEG~\cite{Wang2023EnsembleAnomaly}} & \textbf{SuperAD~\cite{zhang2025superad}} & \textbf{SuperADD (ours)} \\
\midrule
Can          & 0.3 / 0.1 & 0.8 / 0.1 & 17.03 / 9.21 & 1.86 / 0.84 & {\bf 18.14} / {\bf 10.13} & 17.3 / 1.9 & 11.59 / 0.38 \\
Fabric       & 11.5 / 9.8 & 7.6 / 1.0 & 84.95 / 81.76 & 87.46 / 73.37 & 46.53 / 31.74 & 77.4 / 65.3 & {\bf 88.47} / {\bf 86.14} \\
Fruit Jelly  & 8.7 / 8.2 & 20.8 / 18.2 & {\bf 68.1} / {\bf 67.76} & 53.63 / 52.62 & 62.59 / 62.05 & 41.3 / 40.9 & 68.03 / 67.75 \\
Rice         & 3.8 / 4.2 & 15.0 / 0.5 & 66.34 / 66.93 & 63.86 / 63.23 & 72.5 / 73.74 & 60.9 / 61.2 & {\bf 73.83} / {\bf 74.32} \\
Sheet Metal  & 1.8 / 1.1 & 9.3 / 3.8 & 69.69 / 69.74 & {\bf 70.98} / {\bf 70.92} & 64.55 / 62.32 & 59.5 / 59.7 & 55.74 / 55.47 \\
Vial         & 2.3 / 2.2 & 30.5 / 26.5 & 44.01 / 42.12 & 48.73 / 48.83 & 42.67 / 43.96 & 42.8 / 40.8 & {\bf 64.37} / {\bf 63.52} \\
Wallplugs    & 0.0 / 0.0 & 4.4 / 0.3 & 11.99 / 6.24 & 14.38 / 3.40 & 5.5 / 5.06 & 13.7 / 6.7 & {\bf 26.43} / {\bf 18.38} \\
Walnuts      & 1.2 / 1.3 & 34.6 / 13.3 & 68.34 / 67.65 & 67.13 / 58.94 & 67.49 / 67.21 & 69.1 / {\bf 69.1} & {\bf 70.88} / 68.84 \\
\midrule
Mean         & 3.7 / 3.4 & 15.4 / 8.0 & 53.81 / 51.43 & 51.00 / 46.52 & 47.5 / 44.49 & 47.8 / 43.2 & \textbf{57.42} / \textbf{54.35} \\
\bottomrule
\end{tabular}
}
\caption{Performance comparison of segmentation \fone score (in \%) on binarized images for $TEST_{priv}$ / $TEST_{priv,mix}$ set. Bold values indicate the best scores.}
\label{tab:scores_test_private}
\end{table*}

\subsection{Qualitative Results}
\label{subsec:qualitative-results}

\begin{figure*}
    \centering
    \includegraphics[width=1\linewidth]{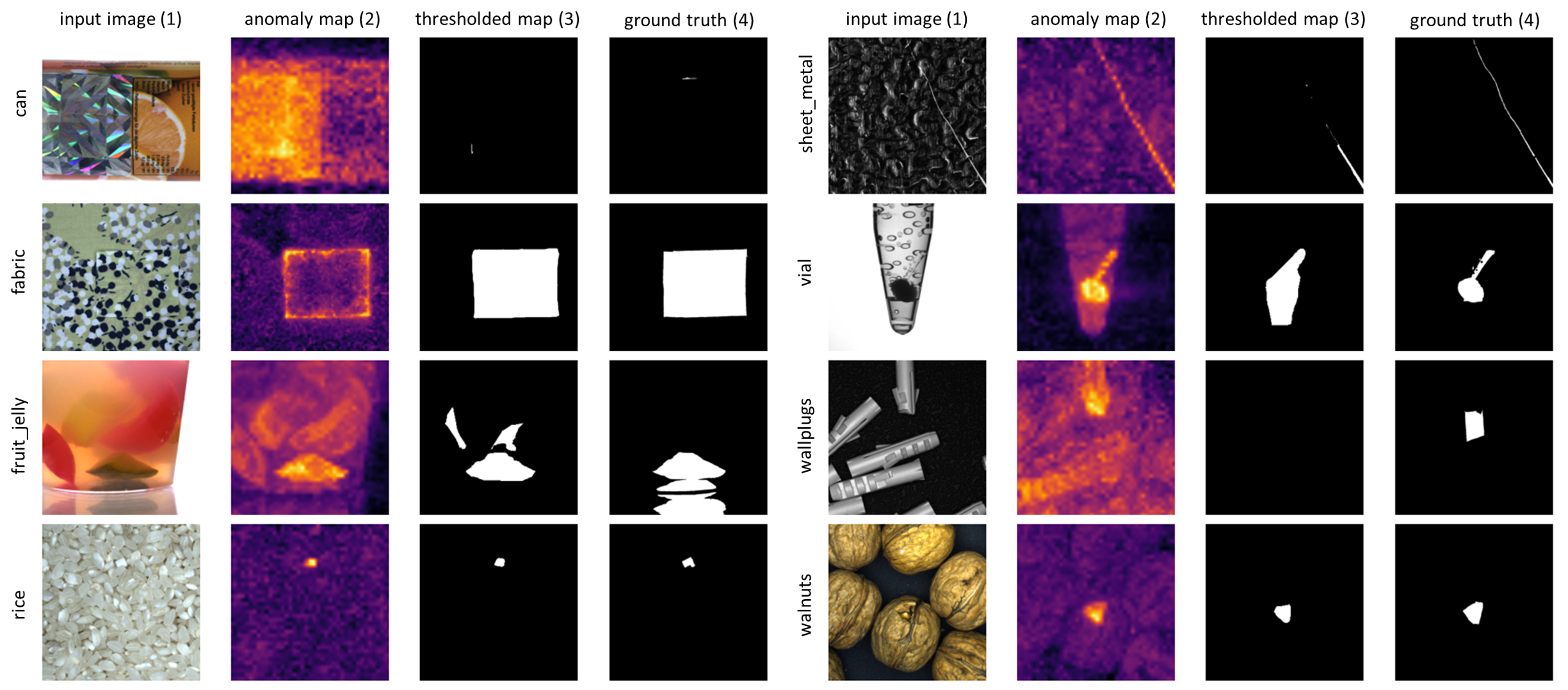}
    \caption{Qualitative results of the proposed method. Columns: (1) the input image, (2) the anomaly map, i.e., the real-valued anomaly scores obtained from the averaged distance maps before post-processing, (3) the binary anomaly map after post-processing, including thresholding, and (4) the ground-truth annotation.}
    \label{fig:Image_Predictions}
\end{figure*}

We provide qualitative results for \testpub in Figure \ref{fig:Image_Predictions}. 
Some presented anomalies are very tiny and need close inspection to be identified including those of categories \textit{can}, \textit{rice}, and \textit{sheet metal}.

For the \textit{sheet metal} example shown in Figure \ref{fig:Image_Predictions}, the model successfully detects the scratch. However, due to a suboptimal threshold, only a portion of it is retained in the thresholded map. The closing operation in the postprocessing ensures that the preserved part of the scratch forms a well-connected region.

An observed limitation of the proposed approach is the detection of missing parts in an image, such as broken pieces in \textit{wallplugs} or an empty \textit{vial}. Similarly, very thin scratches on \textit{can} or \textit{sheet metal} as well as a single hair in \textit{fruit jelly} are often not reliably identified as anomalies.
This drawback, however, does not apply to small defects in general: for categories such as \textit{rice} or \textit{walnut}, the model successfully localizes even very small anomalies.
In contrast, large-scale defects are consistently detected with high accuracy, for example the fabric cutout placed on top of the base material in the \textit{fabric} category, which is clearly and coherently segmented as an anomalous region, owing to the applied post-processing.


%% file: sec/4_discussion_and_conclusion.tex
\section{Discussion}
\label{sec:discussion}

\subsection{Challenges \& Solutions}
\label{subsec:challenges-solutions}

During development, we identified several challenges and adapted our method accordingly.

\textbf{Patch-border artifacts:}
Patch-wise inference with non-overlapping windows led to better overall performance, but at the same time to artifacts at patch borders, including discontinuities in the anomaly map and false detections in regions that were only partially visible within a patch. To mitigate these issues, we introduced overlapping patches in both spatial dimensions. Patch start positions are chosen such that patches are equally distributed and aligned with image borders, guaranteeing a minimum overlap without requiring padding. After inference, redundant predictions in overlapping regions are discarded when reassembling the full-image anomaly map.

\textbf{Disconnected linear anomalies:}
For categories with thin, elongated defects (e.g. scratches on sheet metal), the raw anomaly maps and initial thresholding often produced fragmented segmentations with gaps along the defect. We address this by applying a multi-oriented morphological closing step to the thresholded anomaly map. This connects broken line segments along arbitrary directions while preserving the overall defect shape.

\textbf{Large unfilled regions:}
Following the strategy proposed in SuperAD~\cite{zhang2025superad}, we include a final hole-filling step. After morphological closing, the resulting mask is intersected with a slightly more permissive binary mask obtained by thresholding the anomaly map at $0.8$ times the learned threshold, which constrains the fill operation and reduces over-selection. Finally, we fill closed regions in the binary mask, yielding more coherent anomaly segments and improving robustness to under-segmentation at patch boundaries.

\subsection{Model Robustness \& Adaptability}
\label{subsec:robustness-adaptability}

The model robustness is shown by the performance on the split \testprivmix compared to \testpriv in Table \ref{tab:scores_test_private}. It can be seen that there is a performance drop (overall $\approx 2$ percentage points), most notably for the class can ($\approx 12$ percentage points). But overall, the model still achieved the best performance on the more challenging \testprivmix split with $54.35$\%.

Moreover, the training-free design of our approach contributes to its robustness. Unlike fully trained deep learning models, it does not require potentially unstable optimization procedures, which are prone to convergence failures and often necessitate hyperparameter tuning to establish a reliable training scheme.

\subsection{Future Work}
\label{subsec:future-work}
One promising future direction is to extend this approach to a dual memory bank framework, similar to~\cite{hu2024dmad}. A possible strategy would be to synthesize anomalous images with a diffusion model by transferring defects from MVTec AD 1 onto normal images from MVTec AD 2. The feature alignment strategy of~\cite{xu2025training} could then be used to reduce the gap between synthetic and real defects, enabling the construction of a second, anomaly-focused memory bank.

The two memory banks could be used in a complementary manner: the normal branch would capture deviations from the normal distribution, while the anomalous branch would provide additional evidence of anomaly by matching features to synthetic defect prototypes. This direction remains training-free and could further enhance unsupervised AD performance.

\section{Conclusion}
\label{sec:conclusion}


In this paper, we propose a training-free, class-agnostic method for anomaly segmentation.
Our method builds upon SuperAD~\cite{zhang2025superad}, constructs multi-layer memory banks using DINOv3 as feature extractor, and performs nearest-neighbor-based outlier detection at inference. We introduce a refined subsampling strategy, employ patch-wise image processing, and derive thresholds solely from anomaly-free samples in the training data using the 95\percentile.
We achieved pixel-level \fone scores of $57.42$\% and $54.35$\% for \testpriv and \testprivmix respectively, compared to last years best (ISVL~\cite{wang2025}) with $53.81$\% and $51.43$\% .

Our experiments demonstrate that combining the latest vision foundation model with patch-wise processing, improved subsampling, and post-processing significantly enhances the performance of training-free anomaly detection, achieving generalizable results competitive with state-of-the-art approaches.

